\documentclass[pra,aps,twocolumn,10pt,showpacs]{revtex4-1}
\usepackage{graphicx}

\begin{document}

\title{The Dependence of Frequency Distributions on \\ Multiple Meanings of Words, Codes and Signs}

\author{Xiaoyong Yan$^{1}$}
\author{Petter Minnhagen$^{2}$}
\email{ Petter.Minnhagen@physics.umu.se}

\affiliation{
$^{1}$Institute of Transportation Systems Science and Engineering, Beijing Jiaotong University, Beijing 100044, China\\
$^{2}$IceLab, Department of Physics, Ume{\aa} University, 901 87 Ume{\aa}, Sweden
}

\begin{abstract}

The dependence of the frequency distributions due to multiple meanings of words in a text is investigated by deleting letters. By coding the words with fewer letters the number of meanings per coded word increases. This increase is measured and used as an input in a predictive theory. For a text written in English, the word-frequency distribution is broad and fat-tailed, whereas if the words are only represented by their first letter the distribution becomes exponential. Both distribution are well predicted by the theory, as is the whole sequence obtained by consecutively representing the words by the first $L=6,5,4,3,2,1$ letters. Comparisons of texts written by Chinese characters and the same texts written by letter-codes are made and the similarity of the corresponding frequency-distributions are interpreted as a consequence of the multiple meanings of Chinese characters. This further implies that the difference of the shape for word-frequencies for an English text written by letters and a Chinese text written by Chinese characters is due to the coding and not to the language \emph{per se}.
\end{abstract}

\pacs{89.75.Fb, 89.70-a}

\maketitle

\section{Introduction}
\label{sec1}
Attempts to understand what \textit{linguistic} information is hidden in the \textit{shape} of the word-frequency distribution has a long tradition \cite{estroup16,zipf32,zipf35,zipf49}. A central question in this context is what special principle or property of a language causes the ubiquitous observed "fat tailed' power-law like distribution of word-frequencies\cite{mand53,li92,baayen01,cancho03,mont01,font-clos2013}.

The concept of \textit{randomness} in a text dates back to V. Markov \cite{markov1913, hayes2013}. Markov demonstrated that a text when viewed as a string of letters, contained random features like e.g. how often a randomly chosen letter is followed by a consonant  or a vowel. The concept of randomness in \textit{word-frequency distributions} was emphasized by Simon in Ref. \cite{simon55} who argued that since quite a few completely different systems closely have the same "fat tailed" power-law like frequency distributions, the explanation for this particular shape must be stochastic and independent of any specific information of the language itself. This \textit{Randomness-view} was developed further in a series of paper in terms of concepts like Random Group Formation (RGF), Random Book Transformation and the Meta-book \cite{bern10,bern09,bern11b,baek11,yan15}. According to the "Randomness-view" the shape of the word-frequency distribution is a general consequence of randomness which carries no specific information of the intrinsic structure of a language.\cite{yan16} 

However, even if the frequency distribution of words does not depend on the specifics of the language, it may still depend on how the words are coded by symbols. This is the subject of the present work. We explore the connection between, on the one hand, the shape of the frequency-distribution of the symbols used to represent a written text and, on the other, the information content carried by individual symbols. 

The relation between information content and the shape of a word-frequency distribution goes back to Mandelbrot \cite{mand53}. The focus in this earlier work was the information content obtained by coding an individual word by symbols like individual letters. In the present case we instead start out by taking the written individual words as the symbols and focus on the information loss caused by the fact that an individual written word in a text can have more than one meaning.
In order to investigate this in a systematic way we vary the multiplicity of meanings for a written word by deleting letters. For example \textit{invest}, \textit{inv}, and \textit{i} are the 6-, 3- and 1-first-letter-versions of the full word \textit{investigate}. 

The paper is organized as follows: first we in section \ref{sec2} define the $L$-letter coding model. The multiplicities of words and the corresponding word-frequencies based on the novel Moby Dick by Herman Melville are measured for the $L$-letter word-versions of the text. This directly leads to the question of how the frequency-distributions and multiplicities are connected. Section \ref{sec3} uses the maximum entropy estimate given by Random Group Formation(RGF)-formulation \cite{ baek11,yan15,yan15b,yan16} to obtain such a direct link. This is possible because the RGF-estimate predicts the shape of the frequency distribution using the  multiplicities of meanings as a direct input \cite{yan15,yan16}.

In the light of these findings we in section \ref{sec4} investigate the frequency distribution of Chinese characters for Moby Dick written by Chinese characters. It is found that the character-frequency distribution is very similar to the $L$=3-coding of Moby Dick in English. This suggests that the multiple meanings of Chinese characters are similar to the multiple meanings of the $L$=3-codes. This is in accordance with the findings of Ref.\cite{yan15}. It is also noticed that the coding of a word in English by a three-two-one letter sequence, such that \textit{investigate} is coded by the four symbols \textit{inv, est, iga, te}, leads to an even closer similarity.

Finally section \ref{sec5} contains a summary. An analysis of a second novel (Tess of the D'Urbervilles by Thomas Hardy) is given in an Appendix, as a verification of analysis based on the novel Moby Dick by Herman Melville.

\section{Multiplicity and $L$-letter Coding}
\label{sec2}

\begin{figure}
\centering
\includegraphics[width=0.5\textwidth]{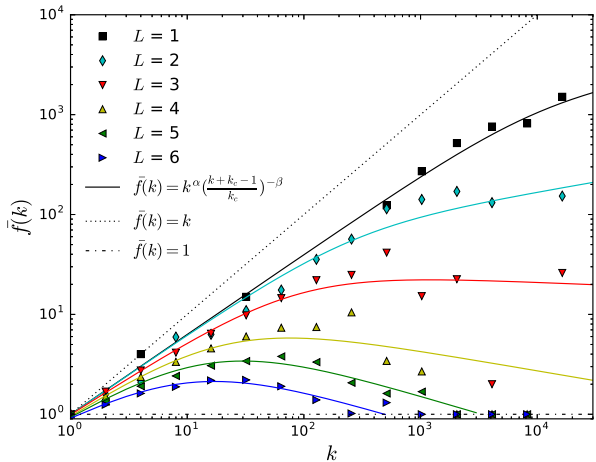}
\caption{The average multiplicity of meanings for different code-lengths $L$. The data are for the code-lengths $L=1-6$. For $L=1$ the data approximately follows the straight line $\overline{f}(k)\approx k^{0.8}$. More generally the data bends for larger $k$ and can be approximated by the function $\overline{f}=k^\alpha ((k+k_c-1)/k_c)^{-\beta}$ (full curves in the figure), as described in the text . The curves are data-fits to this functional form. This a priori knowledge of the multiple meanings will serve as an input in the RGF-prediction.The horizontal dashed-dotted line corresponds to the case when codes only have single meanings, $\overline{f}(k)=1$.The dotted line corresponds to the special case when the codes increases linearly with $k$, $\overline{f}(k)=k$.}      
\label{fig1}
\end{figure}

In an alphabetic text each word is coded by a combination of letters. For example the first word in the novel Moby Dick by Herman Melville is, when written in English, \textit{call}. Thus \textit{call} is the letter-code, or more generally the symbol, for the word and the letter-codes for different words are separated by blanks. In principle different words can sometimes be represented by the same letter-code. This means that a letter-code can represent a word with more than one meaning in the text. The present investigation is addressing the frequency distribution of codes with multiple meanings. The number of words with  multiple meanings within an English novel coded by the English alphabetic letter-code are few  and can to a first approximation be ignored \cite{yan15,yan15b}. In order to systematically investigate the effect of multiplicity we 
instead use a reduced alphabetic letter-code, the $L$-letter representation. In this representation a word is represented by only the first $L$ letters in the English letter-code. Thus for $L=6$ \textit{represented} becomes \textit{repres} and \textit{call} remains \textit{call}, whereas for $L=3$ \textit{represented}
becomes \textit{rep} and \textit{call} becomes \textit{cal}. In the most extreme case $L=1$ \textit{represented} becomes \textit{r} and \textit{call} becomes \textit{c}. The point is that the smaller the value $L$ the harder becomes the interpretation of the text, because the loss of information caused by shortening the letter-code. This missing information has to be supplied by the reader and the actually amount of the information loss is directly related to choosing between the possible multiple meanings the codes have in the text. 

\begin{figure*}
\centering
\includegraphics[width=0.9\textwidth]{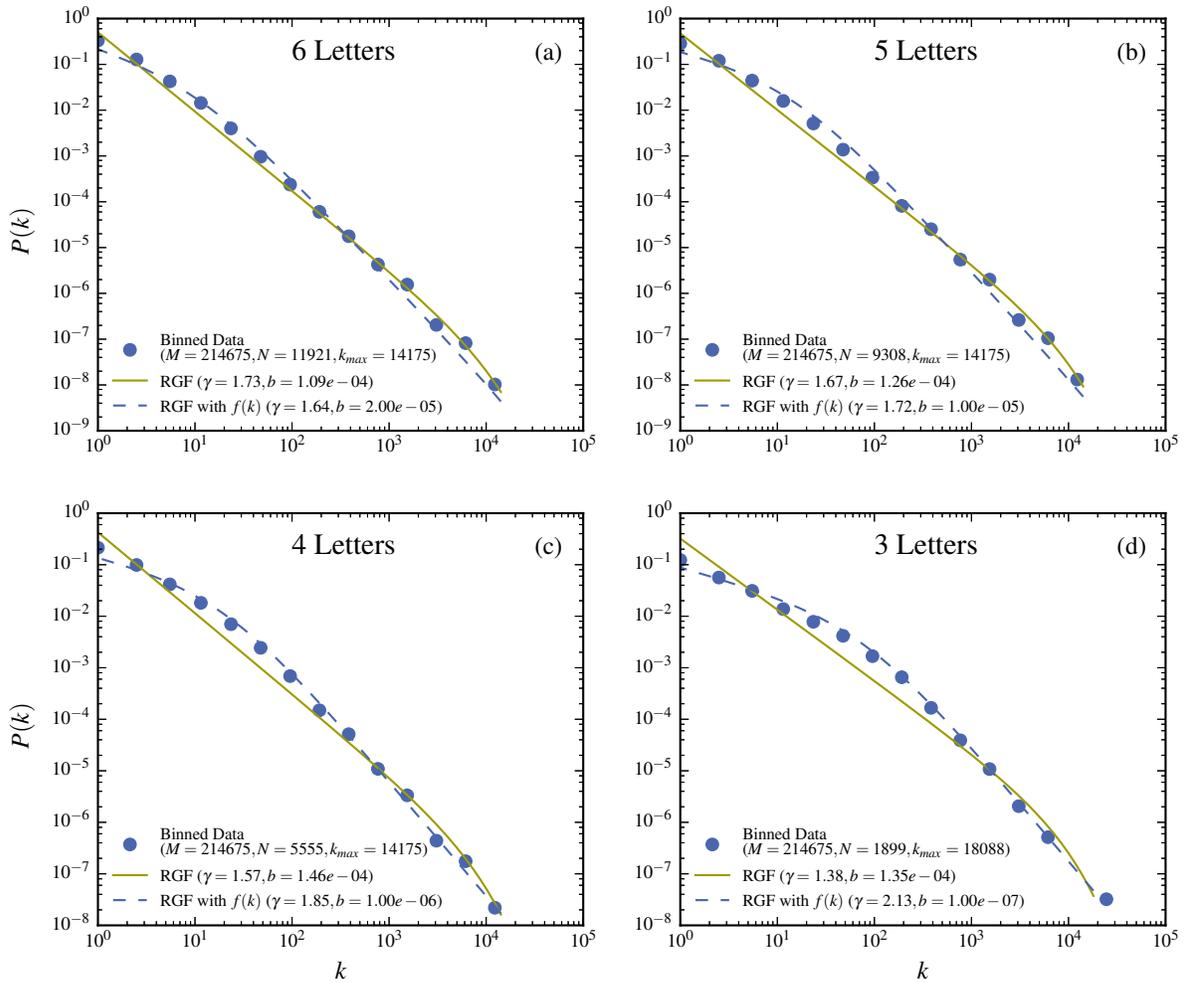}
\caption{Frequency distributions $P(k)$ for the codes $L=6-3$. The data in binned form are given by the filled 
dots. The question is how these distributions are related to the multiple meanings given in Fig. \ref{fig1}. As described in section \ref{sec3}, the RGF predicts $P(k)$ if one a priori knows the triple $(M,N,k_{max})$ and the average multiplicity $\overline{f}(k)$. The a priori known values $(M,N,k_{max})$ are given in the panels. The predictions are given by the dashed curves. The agreements between data and predictions are striking. For comparison the predictions, when assuming that the codes only have single meanings, are given by the full curves. This means that the discrepancies between the full curves and the data are caused by the multiple meanings of the codes.}
\label{fig2}
\end{figure*}

\begin{figure*}
\centering
\includegraphics[width=0.9\textwidth]{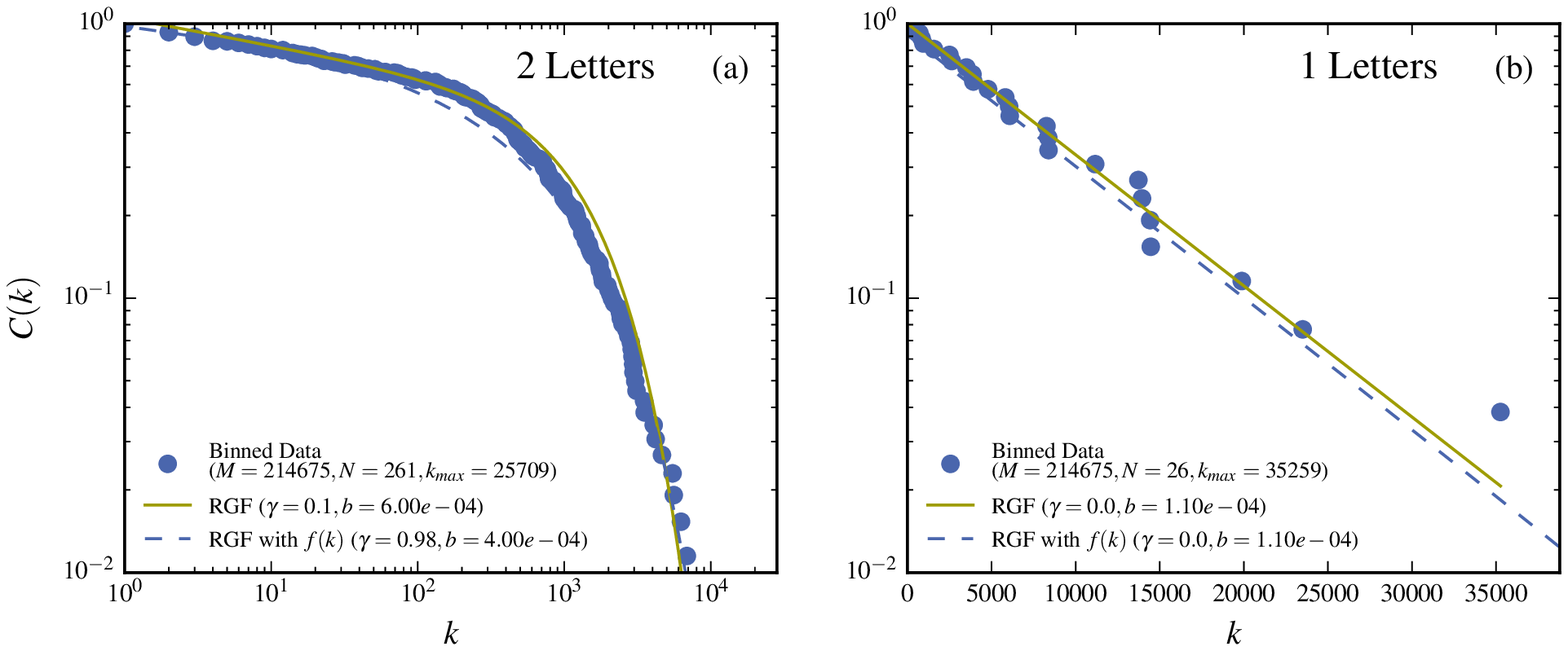}
\caption{Frequency distributions in the cumulant form $P(k>)$ for the code-lengths $L=1$ and $L=2$.The data (without binning) are given by the filled dots. The dashed curves are the RGF-predictions including the a priori knowledge of the multiple meanings given by Fig. \ref{fig1}. The agreements between data and predictions are very good.
The fulled curves are the RGF-predictions assuming that each code only have a single meaning. One notes that the difference between the full and dashed curves is small and that both describe the data very well. The reason, as explained in the text, is that the closer the average multiple meanings are described by 
$\overline{f}(k)=k^\alpha$ the smaller becomes the difference between the dashed and full curves (compare Fig. \ref{fig1}). Also note that Fig. \ref{fig3}(b) is plotted in log-lin which means that $P(>k)$ is close to an exponential in this case.}
\label{fig3}
\end{figure*}

Figure \ref{fig1} shows the average number of meanings, $\overline{f}(k)$, for $L$-codes which occurs $k$ times in the text. An English text of the size of a novel contains few words with multiple meanings within the novel. Hence the multiple meanings of a L-code is obtained from the number of different words which the particular L-code represents within the text. Note that a code which only occurs ones in the text can only have one meaning, so that all curves start from $\overline{f}(k=1)=1$. Also note that the general trend is that $\overline{f}(k)$ increases with decreasing $L$ for a fixed occurrence $k$. This just means that the shorter the $L$-code is for a given $k$ the more meanings has a code with this occurrence on the average. However, apart from this, the $\overline{f}(k)$-curves for different $L$ are non-trivial and we will, in order to facilitate calculations, use the simple parametrization 
\begin{equation}
\label{eq:f(k)}
\overline{f}(k)=k^\alpha \left( \frac{k+k_c-1}{k_c} \right)^{-\beta}
\end{equation}
where $\alpha,\beta$ and $k_c$ are free parameters . This parametrization catches the essential features of the average multiple-meaning function $\overline{f}(k)$ for different $L$-codes (compare Fig. \ref{fig1}). Note that the explicit form of the parametrization given by Eq.(1) has no significance \emph{per se} as long as it catches the essential trend of the data.

The question addressed in the present work is how this the average multiple-meaning function $\overline{f}(k)$ for different $L$-codes
 is reflected in the actual frequency distribution of the corresponding $L$-codes. Figures \ref{fig2} and \ref{fig3} gives the corresponding frequency distributions for the $L$-codes. Figure \ref{fig2} show the frequency probability distributions, $P(k)$, for the $L$-code cases $L=6,5,4,3$. In the case of $L=1$ and 2 the number of different codes are too few (only the 26 letters in the English alphabet in case of $L=1$) to make a binning meaningful and as a consequence the cumulant distributions $P(>k)$ give better representations. Fig. \ref{fig3} gives the  cumulant distributions $P(>k)$ corresponding to $L=1$ and 2. 

 In the following section we show that $\overline{f}(k)$ given by Fig. \ref{fig1} can be explicitly linked to the distributions given in Figs  \ref{fig2} and  \ref{fig3}.

\section{A Direct Link between Multiple Meanings and Frequency Distributions}
\label{sec3}

In Ref. \cite{yan15} it was argued that maximum entropy within the RGF-formulation \cite{baek11} provides a link between the multiple-meanings and the frequency-distribution. The theoretical underpinning for this connection has been further developed in Ref. \cite{yan16}. The present work goes one step further and shows that  such a link is open to quantitative testing.

The RGF-formulation of maximum entropy is based on the information content \cite{baek11,yan16}. It starts out with a random group sorting based on the assumption that each of a set of $M$ objects has a unique label $i$ where $i\in[1,2,..M]$ \cite{baek11,yan15}. Suppose that the number of groups with $k$ objects is $N(k)$, then the total number of objects in these groups are $kN(k)$ and the information needed to localize one of them is $\ln[kN(k)]$ (in nats=natural logarithms). The average of this information over the various group sizes $k$ is $\sum_k [N(k)/N] \ln[kN(k)]$. The group size distribution $P(k)=N(k)/N$ is normalized such that $\sum_k N(k)N=1$ and within the RGF approach this is just the probability distribution for the group sizes. This means that the average information for finding an object is a functional of the distribution $P(k)$ and up to a constant given by $I[P(k)]=\sum_kP(k)\ln[kP(k)]$. The maximum entropy corresponds to the minimum of the functional $I[P(k)]$ \cite{baek11}. The RGF-approach minimizes $I[P(k)]$ under three constraints: fixed normalization ($\sum_kP(k)=1$), fixed average $M/N$, and fixed entropy $S_0$. These constraints are handled with three Lagrangian multiplier and leads to the unique funtional form 
\begin{equation}
P(k)=A \exp(-bk)/k^\gamma
\label{equation:P}
\end{equation}
where $A$, $b$ and $\gamma$ are three constants stemming from the three Lagrangian multipliers. Since the functional form is unique, any sufficient \textit{ a priori} knowledge directly expressible in $P(k)$ can be used to determine the constants. The RGF-description uses the three values $[M,N, k_{max}]$: the number of objects in the largest group, $k_{max}$, is related to $P(k)$ by the relations $\sum_{k=k_c}^MP(k)=1/M$ and $<k_{max}>=\sum_{k=k_c}^MkP(k)$ where the first determines a lower bound on the interval $[k_c,M]$ which contains the largest group and the second the average size of a group within this interval. The actual largest group size $k_{max}$ is used as an input for this average. 

One of the assumption for RGF is that the objects can be uniquely labeled. The question addressed in the present work is how the RGF-form will change if the same label is used for many objects. This means that the information to localize an object belonging to a group size $\ln[kN(k)]$ is lacking by some amount $\ln f(k)$. This amount is what has to be supplied externally in order to uniquely identify the objects. Thus the information available \textit{within} the system is now instead $\ln[kN(k)]-\ln f(k)=\ln[kN(k)/f(k)]$. The average information $I[P(k)]$ changes to $I_f[P(k)]=\sum_kP(k)\ln[kP(k)/f(k)]$ and the corresponding RGF-distribution changes to 
\begin{equation}
P(k)=A \exp(-bk)/(k/f(k))^\gamma
\label{equation:Pf}
\end{equation}
This means that \textit{if} the functional form of the information-loss function $f(k)$ is known, \textit{then} the distribution $P(k)$ can again be predicted from the knowledge of the triple $[M,N, k_{max}]$. 

Note that the labeled entities which are sorted into groups of size $k$ are now themselves subgroups $n_i$ where $i$ labels the subgroups and $n_i$ is the number of objects in the subgroup $i$. The information required to identify an object which belongs to one of the groups which contain $k$ subgroups is hence $\texttt{I}=\ln\left[\sum_{i_k}n_{i_k}\right]$ where $i_k$ numerates all subgroups belonging to groups of size $k$. In the case that $n_{i_k}=1$ for all $i_k$ this reduces to $\texttt{I}_1=\ln [kN(k)]$ where $N(k)$ is the number of groups and each group contains $k$ objects. Consequently the info-loss $\ln f(k)=\texttt{I}-\texttt{I}_1$ is given by 
\begin{equation}
\ln f(k)=\ln\sum_{i_k}\frac{n_{i_k}}{kN(k)}
\end{equation}
or
\begin{equation}
f(k)=\frac{1}{kN(k)}\sum_{i_k}n_{i_k}
\end{equation}
which means that $f(k)$ is the average number of objects per sorted entity. In the specific example of words, this translates to the average number of meanings for a word which occur $k$ times in the text. Thus $\overline{f}(k)$ obtained in Fig. \ref{fig1} are equal to $f(k)$ provided the tiny multiplicity of words within the original English version of Moby Dick can be ignored. 

\begin{figure}
\centering
\includegraphics[width=0.5\textwidth]{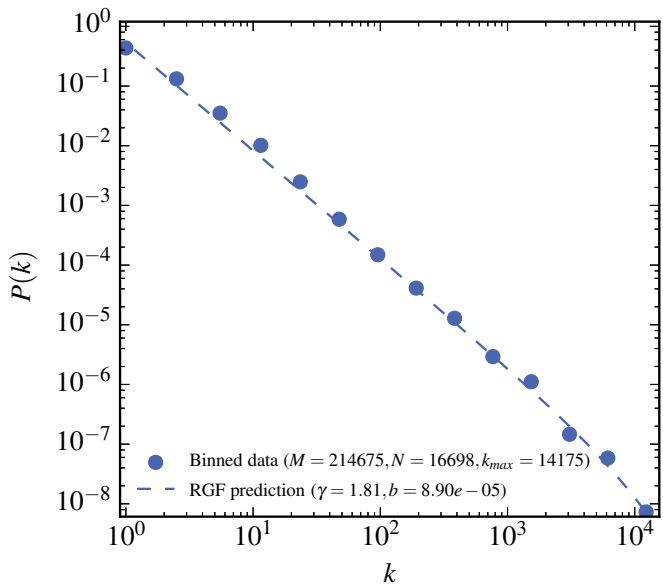}
\caption{Word-frequency distribution for Moby Dick. The RGF-prediction illustrates that the assumption that the words to large extent only carries single meanings gives an excellent prediction provided the words are fully written out: The predicion is completely determined from the a priori knowledge of the values of $(M,N, k_{max})$ and the single meaning assumption $f(k)=1$: Data; filled dot and RGF-prediction; dashed curve.
}      
\label{fig4}
\end{figure}

The statement, that the multiple meaning of words in the original version of Moby Dick can be ignored for practical purposes, corresponds to the assumption that $f(k)=1$ for all $k$. In this case the word frequency is predicted by Eq. (\ref{equation:P}) for the given known values of $[M,N, k_{max}]$. Fig. \ref{fig4} shows that this is indeed the case. Furthermore in Refs. \cite{yan15,yan15b} it was shown that the single meaning assumption $f(k)=1$ in general gives very good frequency distribution predictions for texts written by normal letter-alphabets, suggesting that the multiple meanings per written word is small and often can be ignored.

In Fig. \ref{fig2} the assumption of single meanings $f(k)=1$ is tested on the $L$-letter codes with $L=6,5,4,3$. The predictions from the RGF-estimate correspond to the full  curves in the figure. For $L=6$ the deviation between the data and the prediction is rather small. However, as $L$ decreases the deviation becomes larger and for $L=3$ it is substantial. In order to test if this deviation is due to the multiple meanings of the $L$-letter codes, all one has to do is to use the actual known multiplicity $f(k)=\overline{f}(k)$ in the RGF-estimate instead of $f(k)=1$ (compare Fig. \ref{fig1}). This changes the RGF-prediction to the dashed curves in Fig. \ref{fig2}. The agreement between the predictions (dashed curves) and the data (dots) is striking. Thus the RGF-estimate provides a direct quantitative link between the multiplicity of meanings and the corresponding frequency distributions. 
\begin{figure}
\centering
\includegraphics[width=0.5\textwidth]{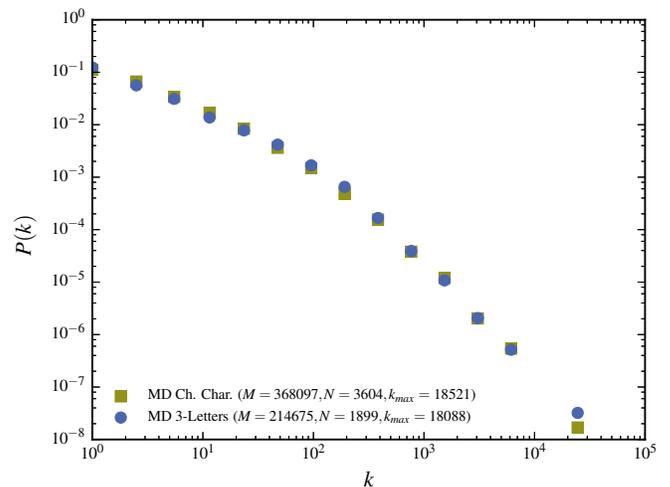}
\caption{ Direct comparison between $L=3$-coding of Moby Dick in English and Moby Dick in written Chinese by Chinese characters. Note the strong overlap. In case of the $L=3$-coding the shape of the frequency distribution is directly connected to the multiple meanings of the codes (compare discussion of Fig. \ref{fig1}). This suggest to us that the multiple meanings of Chinese characters is likewise link to the multiple meanings of the Chinese characters.}
\label{fig5}
\end{figure}

Fig. \ref{fig3} gives the cases for $L=1$ and 2. The most striking case is $L=1$, which corresponds to representing each word in the text by just its first letter. As seen from Fig. \ref{fig1}, the $\log \overline{f}(k)$  versus $\log k$ is roughly linear, so that in this case $\overline{f}(k)=k^\alpha$ with $\alpha \approx 0.8$. This means that 
Eq.(\ref{equation:Pf} reduces to the form Eq.(\ref{equation:P}). Consequently the full prediction with the known $\overline{f}(k)$ and the one assuming $\overline{f}(k)=1$ should in this case yield approximately the same prediction. As seen in Fig. \ref{fig3}(b) this is indeed the case: both predictions describe the data very well. Note that Fig. \ref{fig3}(b) is plotted is in lin-log so that the fact that the data to good approximation falls an a straight line shows that the corresponding cumulant $P(>k)$, and as a consequence also $P(k)$, are close to exponential. It is interesting to note that if $\overline{f}(k)=k$, then the RGF-prediction given by Eq.(\ref{equation:Pf}) reduces to just an exponential. The distribution of first letters in Moby Dick approaches this limiting case (compare Fig. \ref{fig1}).  

One may also note that the RGF-estimate is not restricted to broad power-law like distributions, as illustrated by the cases $L=1$ and $L=2$. This feature has been further explored in Ref. \cite{yan16}.

\begin{figure*}
\centering
\includegraphics[width=1\textwidth]{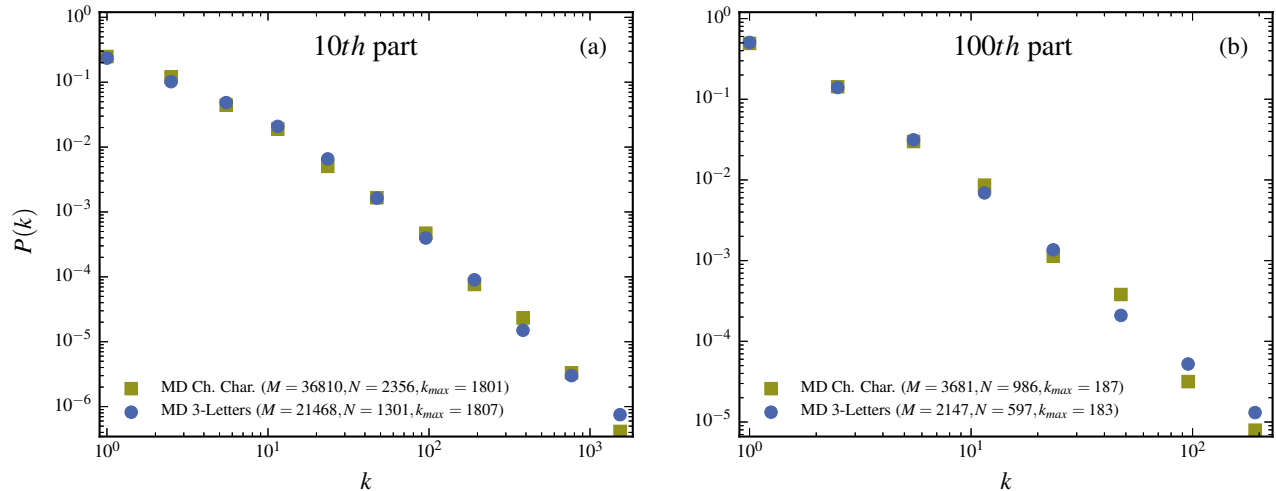}
\caption{ Direct comparison between $L=3$-coding of Moby Dick in English and Moby Dick in written Chinese by Chinese characters when taking  parts of the texts. The multiple meanings of the $L=3$-codes decrease when shortening the text and this changes the frequency distribution.Provided the shape of the frequency distribution for Chinese characters is likewise linked to the multiplicity of meanings, one expects a very similar change in the distributions. Fig. \ref{fig6} (a) and (b) shows that this expectation is indeed borne out. The comparisons are for a $10^{th}$-part (a) and a $100^{th}$-part (b).
}      
\label{fig6}
\end{figure*}

\section{Qualitative connection to Chinese Characters}
\label{sec4}
In Ref. \cite{yan15} it was argued, that the deviation between the RGF-estimate with $f(k)=1$ for a text written by Chinese characters and the frequency distribution of the characters, was caused by the multiple meanings of the Chinese characters. However, in the case of Chinese characters there is no easy way to directly obtain the multiple meanings of characters and hence the function corresponding to $\overline{f}(k)$. Thus a direct link, like the one obtained for the $L$-letter codes, is harder to obtain. However, qualitatively $\overline{f}(k)$ corresponding to Chinese characters has to be a function starting from $f(k=1)=1$ (because a character which only occurs once in the text can only have one meaning), then it increases with $k$, but with some cut-off because even the most common character does have a limited number of meanings. Thus you expect something qualitative similar to the multiple meanings in Fig. \ref{fig1}. To test this qualitative similarity, Fig. \ref{fig5} compares the frequency distribution of $L$=3-letter codes for Moby Dick with the character distribution of Moby Dick in Chinese translation. In other words you compare the same text representated by two completely different symbolic systems. As seen in Fig. \ref{fig5} the two corresponding frequency distribution are very similar which suggests that the $f(k)$ corresponding to Moby Dick in Chinese characters are also akin to the one for the $L$=3-letter code given in Fig. \ref{fig1}. This similarity is further enforced by comparing parts of Moby Dick for the $L$=3-letter codes and the Chinese characters. Fig. \ref{fig6} compares the $10^{th}$-parts and $100^{th}$-parts and again the close similarity remains. In the case of the $L$=3-letter codes, we know that the shape of the distribution is directly linked to the multiplicity of meanings. The close similarity with the same text-parts written by Chinese characters, suggest to us that the change in shape is also in the case Chinese characters linked to the multiplicity of meanings.  

\begin{figure*}
\centering
\includegraphics[width=1\textwidth]{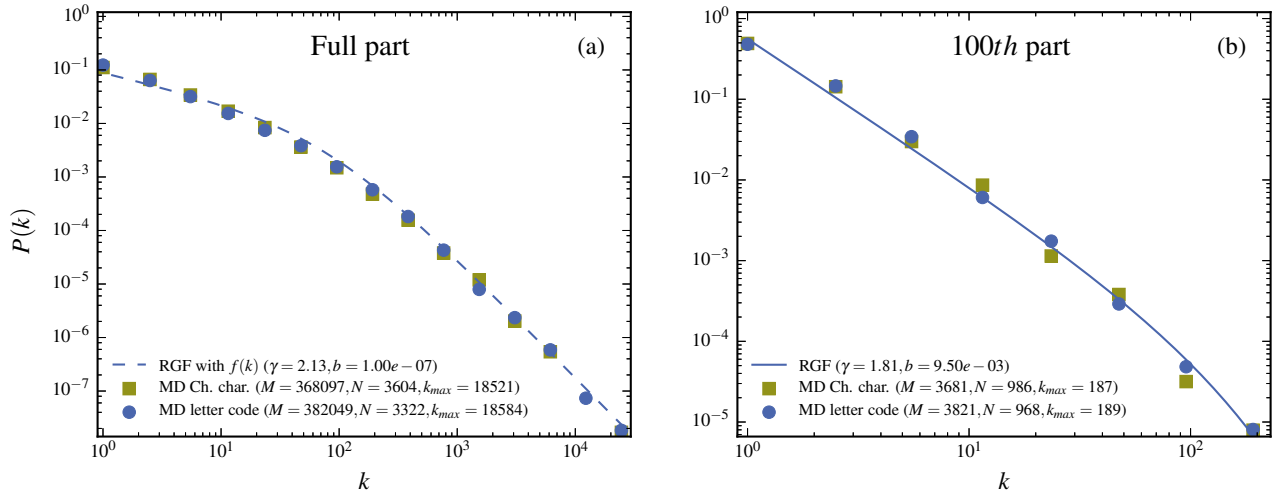}
\caption{Comparison between Mody Dick represented by Chinese characters and Moby Dick represented by the partial letter code described in the text. These two representation of the same text have closely the same total number of codes, $M$, number of different codes, $N$, and the number of repetitions of the most frequent code $k_{max}$. Fig. \ref{fig7}(a) illustrates that also the corresponding frequency distributions are closely the same. In addition Fig. \ref{fig7}(b) illustrates that both frequency distributions changes in the same way, when shortening the text: the $100^{th}$-parts of the two different codes are also closely similar. For the $100^{th}$-parts the RGF-prediction, assuming single meanings ($\overline{f}(k)=1$), gives a good prediction for both codes (full curve, the difference of triples $(M,N,k_{max}$ are so small that the two RGF-predictions overlap). However, RGF-prediction assuming single meaning differs from the data in Fig. \ref{fig7}(a). Inserting the known $\overline{f}(k)$ for the $L=3$-case (compare Fig. \ref{fig1} and Fig. \ref{fig2}(d)) into the RGF-prediction gives the dashed line in Fig. \ref{fig7}(a). This suggests that multiple meanings are crucial for the shape of the frequency distribution also in case of Chinese characters.  
}      
\label{fig7}
\end{figure*}
The special thing  with the $L$-letter coding is that for this coding the multiplicity of meanings for the codes can be easily obtained. More generally you can, of course, device  other partial letter codes. However for such codes it is often harder to directly extract the multiplicity of meanings. Fig. \ref{fig7} shows the frequency distribution for Moby Dick using the following partial letter-code: an English word is coded by three-two-one letter sequences, such that \textit{sequence} is coded by the three symbols \textit{seq, uen, ce,}. Represented in this way Moby Dick consists of  $M=382049$ codes of which $N=3322$ are different and the most common code appears $k_{max}=18584$ times in the text. Moby Dick in Chinese written by Chinese characters contains $M=368097$ characters of which $N=3604$ are different Chinese characters and the most common appears $k_{max}=18521$ times. The point is that the triple $(M,N,k_{max})$ is approximately the same for these two cases. RGF predicts the frequency distribution from this together with $f(k)$. The results are given in Fig. \ref{fig7}: Fig. \ref{fig7}(a) is for the full Moby Dick written in the above partial letter code, whereas Fig. \ref{fig7}(b) is for the $100^{th}$-parts. As expected the RGF prediction for $f(k)=1$ gives a good agreement for the $100^{th}$-part because the multiple meanings for the codes and characters are almost negligible for text sufficiently short text. For the full text there is, on the other hand, a discrepancy, which in Fig. \ref{fig2}(d) was attributed to the multiple meanings of the codes and characters for a longer text. If we assume that this multiplicity, in case of both the above partial letter code version  and Chinese characters in Fig. \ref{fig7}(a), are very similar to the $L$=3-letter code version and use the corresponding  $L=3$ letter $f(k)=\overline{f}(k)$, the dashed curve in Fig. \ref{fig7} is obtained. The close agreement again suggest that the multiple meanings are directly linked to the shape of the corresponding frequency distributions.

One might more generally ask, that if you code the same text in two difference ways, such that the total number of codes $M$, the number of different codes $N$ and the occurrence of the most frequent code $k_{max}$ are the same, would that also imply that the $f(k)$ has to be rather similar because after all the information content in the total text is also closely the same. Fig. \ref{fig7} suggest that this might be the case.

\section{Summary}
\label{sec5}

The relation between multiple meanings and the shape of frequency distributions were explored by using a particular letter-coding of words in a text from which the multiple meanings of the codes could be extracted. By using the maximum entropy principle in the RGF information-based formulation together with the known multiplicity as an input, it was demonstrated that the corresponding frequency distributions are predicted to very good approximation. 

From this we concluded that the shape of the frequency distribution essentially is determined by how the text is coded. More precisely we concluded that the shape of the frequency distribution is to good approximation obtained from the maximum entropy principle, provided one knows the total number of symbols M, the number of specific symbols, N, and the occurrence of the most frequent symbol, $k_{max}$, together with the average multiplicity of the symbols, $f(k)$. Thus knowledge of the frequency of the most common symbol and the average multiplicity of the symbols together with the total number of symbols and the number of specific symbols is basically the only information which is reflected in the frequency distribution. Or, expressed in another way, the frequency distribution for words or characters of a written language carries basically no additional specific information about the underlying language.
More language specific properties are instead reflected in correlations between different words. Nevertheless, as shown in Ref.\cite{bern09}, the triple $(M,N, k_{max})$ by itself is on the average different for texts written by different authors and can be used as a "author-fingerprint".

	From the similarity of a text expressed by a partial letter code and by Chinese characters, we concluded that the Chinese characters are symbols in the same sense as the partial letter-codes. Thus also for texts written by Chinese characters the frequency distributions are to good approximation determined by just the total number of symbols, the number of specific symbols and the occurrence of the most frequent symbol, together with the average multiplicity.

One may then ask how this conclusion is related to the view that the particular shape of the Chinese character frequency distribution is explicitly related to  particular detailed features of the Chinese character construction like e.g. its hierarchic structure.\cite{deng14} The conclusion drawn from the present work is that the shape of the character distribution is unrelated to such specific features. In a more general context the answer is that for complex systems, the global statistical macroscopic features do often not depend on the microscopic details of the system.\cite{yan16} 

We also note that if a text is coded by unknown symbols, an analysis of the symbol-frequency distribution may give a clue to the average multiple meanings of the symbols. If this analysis suggests almost no multiple meanings, this one implies that each symbol codes for a single word. 

\appendix
\section{ }
The purpose of this Appendix is show that the results found in the present investigation are not special properties of a particular novel. To this end we give the results for the same analysis based on second novel. We have chosen Tess of the D'Urbervilles(TD) by Thomas Hardy which is characterized by the triple $(M,N,k_{max})=(152952, 11917, 8717)$ to be compared to Moby Dick which is characterized by the triple $(M,N,k_{max})=(214675, 16698, 14175)$.

According to RGF these two triples contain enough information for predicting the corresponding word-frequency distributions. The RGF-prediction TD is given in Fig.\ref{figA4} which should be compared to the corresponding prediction for Moby Dick given in Fig.\ref{fig4}. In both cases the agreement between prediction and data is very good. 

From the point of view in the present paper, such an excellent agreement presumes that the effect of multiple meanings of the words can be ignored. If the multiple meanings cannot be ignored then multiple meanings can be approximately taken into account from the average number of meanings for a word which occurs $k$-times in the text, $f(k)$. RGF then predicts the frequency distribution from the knowledge  of $(M,N,k_{max}, f(k))$.

Fig.\ref{figA1} shows the multiple meanings of the L-letter codes for TD and should be compared to corresponding Fig.\ref{fig1} for Moby Dick. The two figures are very similar. 

Fig.\ref{figA2} shows the TD-data for the L-letter codes L=6,5,4,3 together with the RGF prediction ignoring multiple meanings (full drawn curves) and including multiple meanings (dashed curves). The conclusion drawn in the present paper is that the discrepancy between the data and the full drawn curves are caused by the multiple meanings of the L-letter codes. The similarity with the corresponding Fig.\ref{fig2} for Moby Dick is re-assuring. Comparing Fig.\ref{figA3} for TD with Fig.\ref{fig3} for Moby Dick shows that also the two extreme L-letter codes L=1,2 show precisely the same features for both the novels.

Finally as shown in Fig.\ref{figA5}, comparing TD written in English L=3-letter code with TD written in Chinese with Chinese characters gives the same striking overlap as the corresponding data for Moby Dick given in Fig.\ref{fig5}.

\renewcommand\thefigure{A\arabic{figure}}
\setcounter{figure}{0}

\begin{figure*}
\centering
\includegraphics[width=0.5\textwidth]{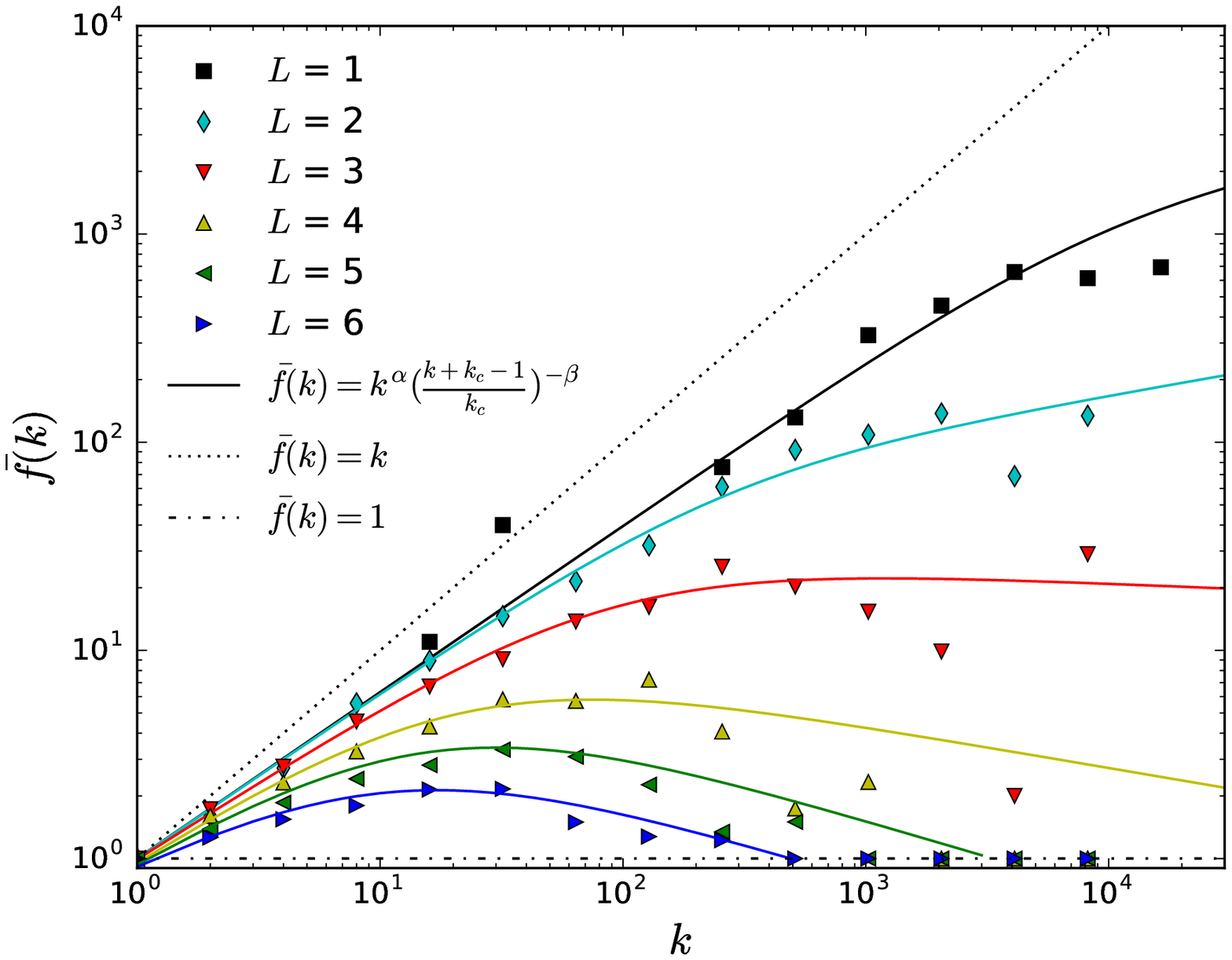}
\caption{The novel TD (Tess of the D'Urbervilles by Thomas Hardy): Average multiplicity of meanings for different code-lengths $L$. Compare Fig.1 which gives the same data for Moby Dick.}      
\label{figA1}
\end{figure*}

\begin{figure*}
\centering
\includegraphics[width=0.9\textwidth]{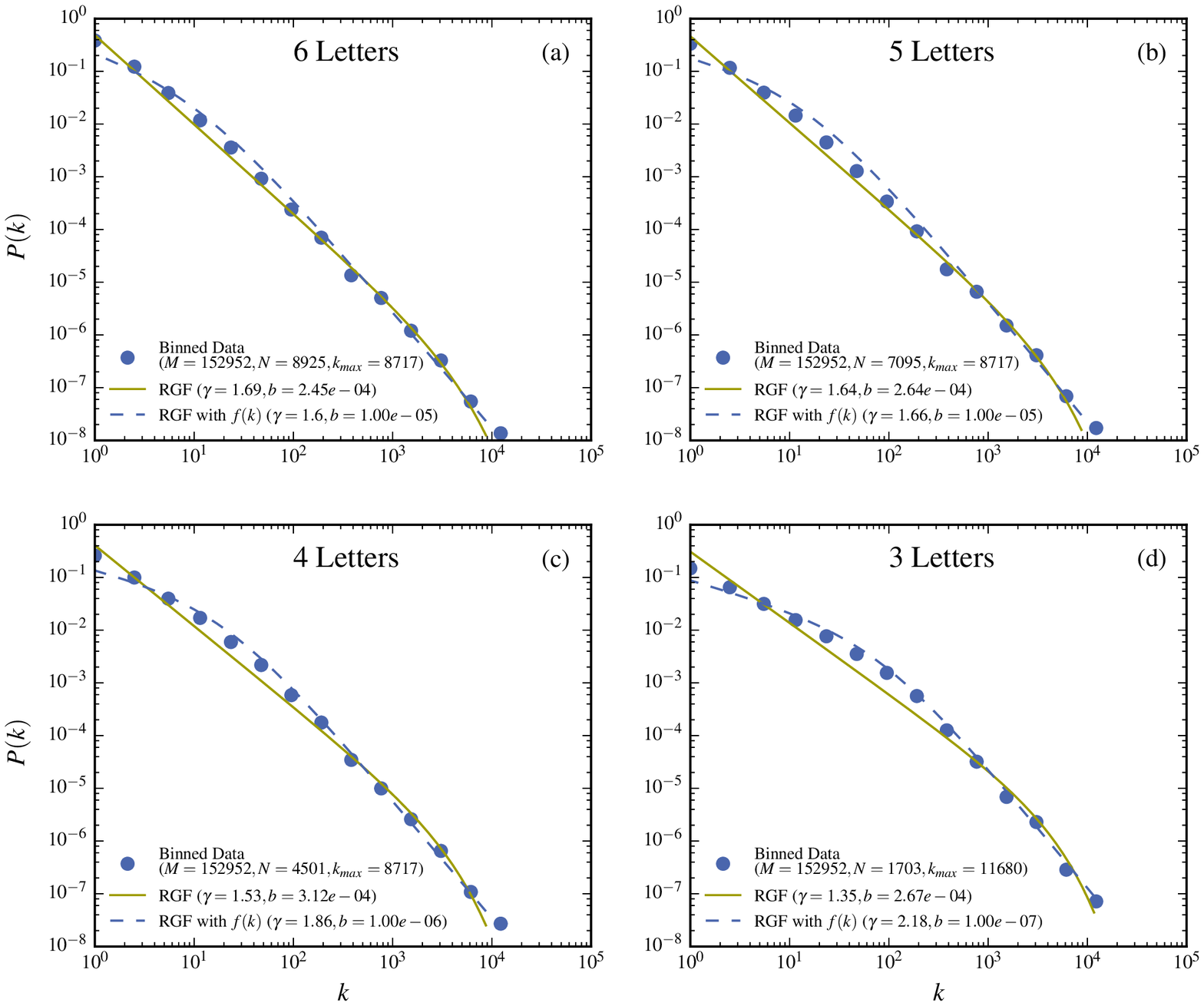}
\caption{The novel TD: Frequency distributions $P(k)$ for the codes $L=6-3$. Compare Fig.2 which gives the same data for Moby Dick.}
\label{figA2}
\end{figure*}

\begin{figure*}
\centering
\includegraphics[width=0.9\textwidth]{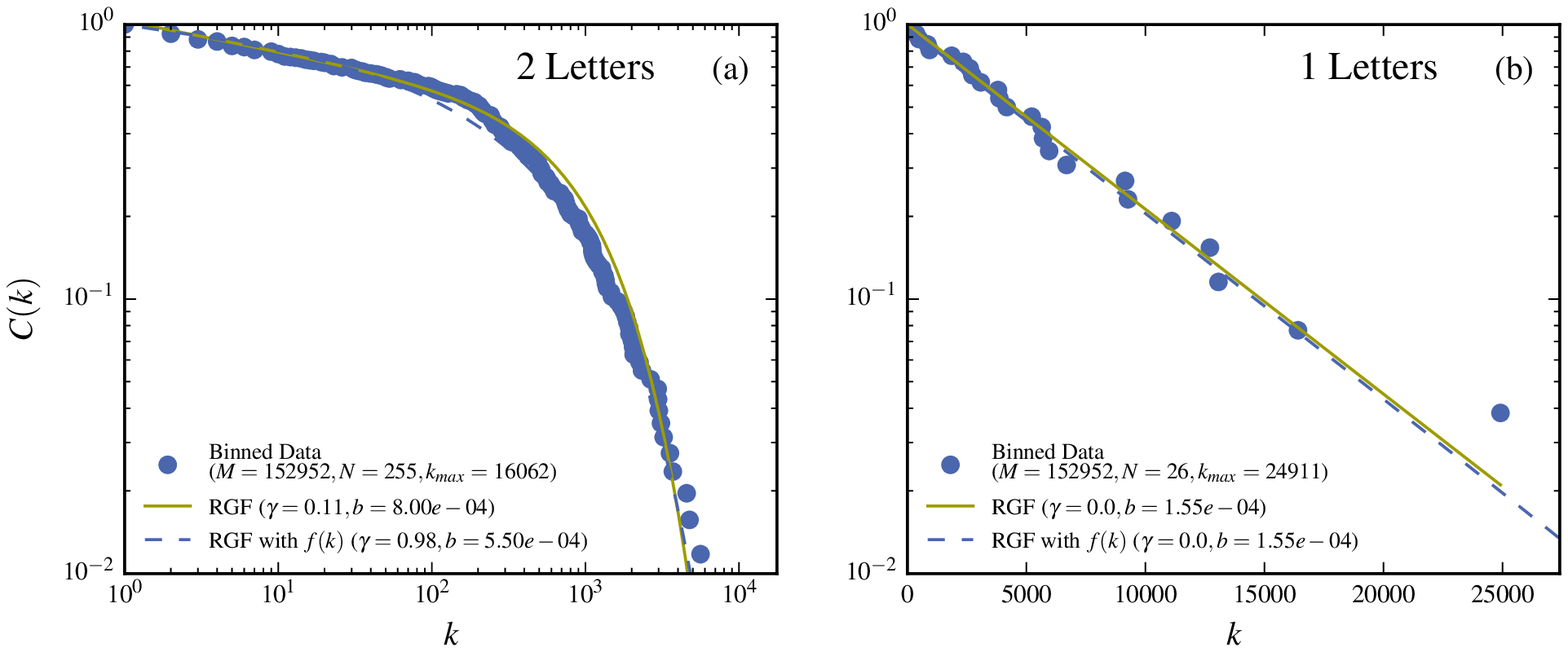}
\caption{The novel TD: Frequency distributions in the cumulant form $P(k>)$ for the code-lengths $L=1$ and $L=2$. Compare Fig.3 which gives the same data for Moby Dick}
\label{figA3}
\end{figure*}

\begin{figure*}[htbp]
\begin{minipage}[t]{0.49\textwidth}
\begin{flushleft}
\includegraphics[width=1\textwidth]{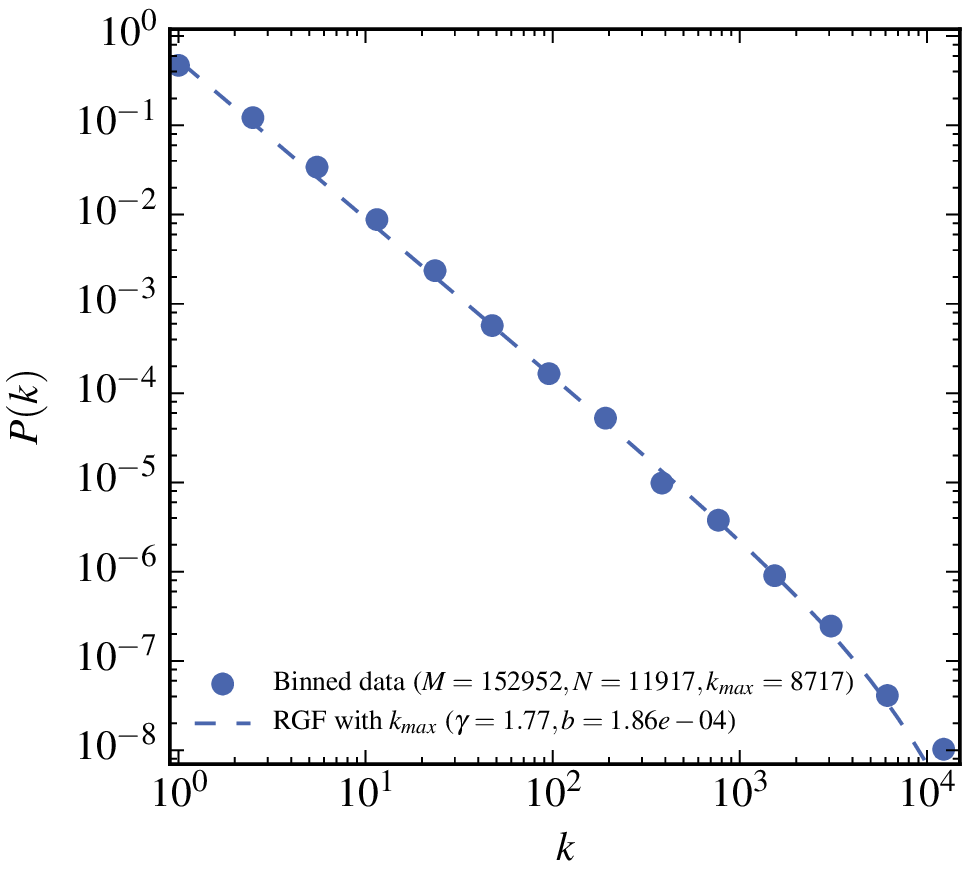}
\caption{The word-frequency distribution for the novel TD. The RGF-prediction illustrates that the assumption that the words to large extent only carries single meanings gives an excellent prediction provided the words are fully written out: The predicion is completely determined from the a priori knowledge of the values of $(M,N, k_{max})$ and the single meaning assumption $f(k)=1$: Data; filled dot and RGF-prediction; dashed curve. Compare Fig.4 which gives the word-frequency distribution for the novel Moby Dick.
}      
\label{figA4}
\end{flushleft}
\end{minipage}
\noindent
\begin{minipage}[t]{0.49\textwidth}
\begin{flushright}
\includegraphics[width=1\textwidth]{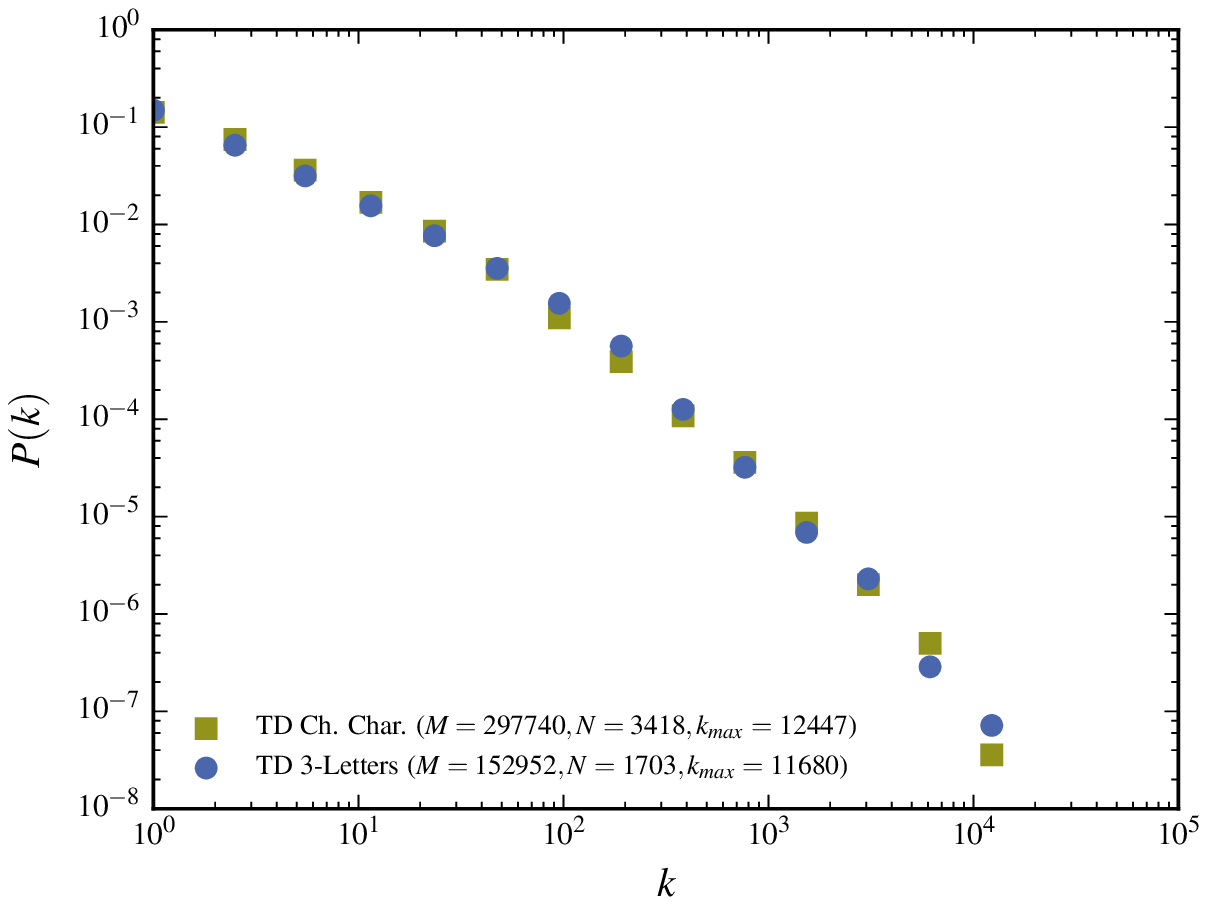}
\caption{ Direct comparison between $L=3$-coding of TD in English and TD in written Chinese by Chinese characters. Note the strong overlap. Compare Fig.5 which gives the same data for Moby Dick. In case of the $L=3$-coding the shape of the frequency distribution is directly connected to the multiple meanings of the codes (compare discussion of Fig. \ref{fig1}). This suggest to us that the multiple meanings of Chinese characters is likewise link to the multiple meanings of the Chinese characters.}
\label{figA5}
\end{flushright}
\end{minipage}
\end{figure*}


\begin{thebibliography}{99}
\bibitem{estroup16}
J.B. Estroup, Les Gammes St\'enographiques, fourth ed., Institut Stenographique de France, Paris, 1916.
\bibitem{zipf32}
G.K. Zipf, Selective Studies of the Principle of Relative Frequency in Language, Harvard University Press, Cambridge, 1932.
\bibitem{zipf35}
G.K. Zipf, The Psycho-Biology of Language: an Introduction to Dynamic Philology, Mifflin, Boston, 1935.
\bibitem{zipf49}
G.K. Zipf, Human Bevavior and the Principle of Least Effort, Addison-Wesley, Reading, 1949.
\bibitem{mand53}
B. Mandelbrot, An Informational Theory of the Statistical Structure of Languages, Butterworth, Woburn, 1953.
\bibitem{li92}
W. Li, Random texts exhibit Zipf's-law-like word frequency distribution, IEEE T. Inform. Theory 38 (1992) 1842-1845.
\bibitem{baayen01}
R.H. Baayen, Word Frequency Distributions, Kluwer Academic, Dordrecht, 2001. 
\bibitem{cancho03}
R.F. i Cancho, R.V. Sol\'e, Least effort and the origins of scaling in human language, Proc. Natl. Acad. Sci. U.S.A. 100 (2003) 788-791.
\bibitem{mont01}
M.A. Montemurro, Beyond the Zipf-Mandelbrot law in quantitative linguistics, Physica A 300 (2001) 567-578.
\bibitem{font-clos2013}
F, Font-Close, G. Boleda, A. Corral, A scaling law beyond Zipf's law and its relation to Heaps' law, New J. Phys 15 (2013) 093033.

\bibitem{markov1913}
A.A. Markov, An example of statistical investigation of the text \emph{ Eugene Onegin} 
concerning the connection of samples in chains, Sci. Context 19 (2006) 595-600.

\bibitem{hayes2013}
B. Hayes, First links in Markov chains, Am. Sci. 101 (2013),92-97.

H. Simon, On a class of skew distribution functions, Biometrika 42 (1955) 425-440.

\bibitem{simon55}
H. Simon, On a class of skew distribution functions, Biometrika 42 (1955) 425-440.

\bibitem{bern10}
S. Bernhardsson, L.E.C. da Rocha, P. Minnhagen, Size dependent word frequencies and the translational invariance of books, Physica A 389 (2010) 330-341.
\bibitem{bern09}
S. Bernhardsson, L.E.C. da Rocha, P. Minnhagen, The meta book and size-dependent properties of written language, New J. Phys. 11 (2009) 123015.
\bibitem{bern11b}
S. Bernhardsson, S.K. Baek, P. Minnhagen, A paradoxical property of the monkey book, J. Stat. Mech. 7 (2011) PO7013.
\bibitem{baek11}
S.K. Baek, S. Bernhardsson, P. Minnhagen, Zipf's law unzipped, New J. Phys. 13 (2011) 043004.
\bibitem{yan15} 
X. Yan, P. Minnhagen, Maximum entropy, word-frequency, Chinese characters, and multiple meanings, PLoS ONE 10 (2015) e0125592.
\bibitem{yan15b}
X. Yan, P. Minnhagen, Randomness versus specifics for word-frequency distributions, Physica A, 444 (2016) 828-837
\bibitem{yan16}
X. Yan, P. Minnhagen, H.J. Jensen , The likely determines the unlikely, Physica A 456 (2016) 112-119.
\bibitem{deng14} W. B. Deng, A.E. Allaverdyan, B.Li, Q.A. Wang, Rank-frequency relation for Chinese characters, Eur. Phys. J. B 87 (2014) 47


\end{thebibliography}
\end{document}